\documentclass[11pt]{article}

\usepackage[final]{acl}
\usepackage{cuted}

 \usepackage{microtype}

\usepackage[english]{babel} % English as the main language
\usepackage{times}
\usepackage{graphicx}
\usepackage{latexsym}
\usepackage{xcolor}
\usepackage{amssymb}
\usepackage{amsthm}
\usepackage{booktabs}
\usepackage{ulem}   
\usepackage{float}
\usepackage{subcaption}
\usepackage{caption}
\usepackage{float}
\usepackage[T1]{fontenc}
\usepackage[utf8]{inputenc}

\usepackage{microtype}
\usepackage{amsmath}
\usepackage{inconsolata}

\title{\textsc{GRPO-LEAD}: A Difficulty-Aware Reinforcement Learning Approach for Concise Mathematical Reasoning in Language Models}

\author{
  Jixiao Zhang\footnotemark[1] \\
  Johns Hopkins University \\
  \texttt{jzhan432@jh.edu}
  \And
  Chunsheng Zuo\footnotemark[1] \\
  Johns Hopkins University \\
  \texttt{czuo3@jh.edu}
}

\begin{document}
\maketitle

% Somewhere after \maketitle (or immediately after \author)
\footnotetext[1]{Equal contribution.}

% \begin{abstract}
% Recent advances in R1-like reasoning models leveraging Group Relative Policy Optimization (GRPO) have significantly improved the performance of language models on mathematical reasoning tasks. However, current GRPO implementations encounter critical challenges, including reward sparsity due to binary accuracy metrics, limited incentives for conciseness, and insufficient focus on complex reasoning tasks. To address these issues, we propose GRPO-LEAD, a suite of novel enhancements tailored for mathematical reasoning. Specifically, GRPO-LEAD introduces (1) a length-dependent accuracy reward to encourage concise and precise solutions, (2) an explicit penalty mechanism for incorrect answers to sharpen decision boundaries, and (3) a difficulty-aware advantage reweighting strategy that amplifies learning signals for challenging problems. Furthermore, we systematically examine the impact of model scale and supervised fine-tuning (SFT) strategies, demonstrating that larger-scale base models and carefully curated datasets significantly enhance reinforcement learning effectiveness. Extensive empirical evaluations and ablation studies confirm that GRPO-LEAD substantially mitigates previous shortcomings, resulting in language models that produce more concise, accurate, and robust reasoning across diverse mathematical tasks. Our source code, generated dataset, and models are available after the acceptance of this paper.
% \end{abstract}
\begin{abstract}
Group Relative Policy Optimization (GRPO), which is widely adopted by R1-like reasoning models, has advanced mathematical reasoning. Nevertheless, GRPO faces challenges in reward sparsity, verbosity, and inadequate focus on problem difficulty. We propose \textsc{GRPO-LEAD}, enhancing GRPO with: (1) length-regularized rewards to encourage conciseness while maintaining accuracy; (2) explicit penalties for incorrect solutions to improve model precision; and (3) difficulty-aware advantage reweighting for robust generalization on challenging problems. Comprehensive evaluations demonstrate that \textsc{GRPO-LEAD} significantly improves reasoning accuracy, conciseness, and efficiency. Our approach achieves state-of-the-art performance for 14B-scale models, underscoring the synergy of our methods with appropriate model scale and high-quality data. Our source code, generated dataset, and models are available at \url{https://github.com/aeroplanepaper/GRPO-LEAD}.
% Recent R1-like reasoning models using Group Relative Policy Optimization (GRPO) have advanced mathematical reasoning, yet GRPO faces limitations like reward sparsity, poor conciseness incentives, and inadequate focus on problem difficulty. We propose GRPO-LEAD, enhancing GRPO with: (1) length-regularized rewards to encourage conciseness while maintaining accuracy; (2) explicit penalties for incorrect solutions to improve model precision; and (3) difficulty-aware advantage reweighting for robust generalization on challenging problems. Comprehensive evaluations demonstrate GRPO-LEAD significantly improves reasoning accuracy, conciseness, and efficiency. Our approach achieves state-of-the-art performance for 14B-scale models, underscoring the synergy of our methods with appropriate model scale and high-quality data. Our source code, generated dataset, and models are available after the acceptance of this paper.
\end{abstract}

\section{Introduction}
\begin{figure*} 

     \centering 

     \includegraphics[width=1\linewidth]{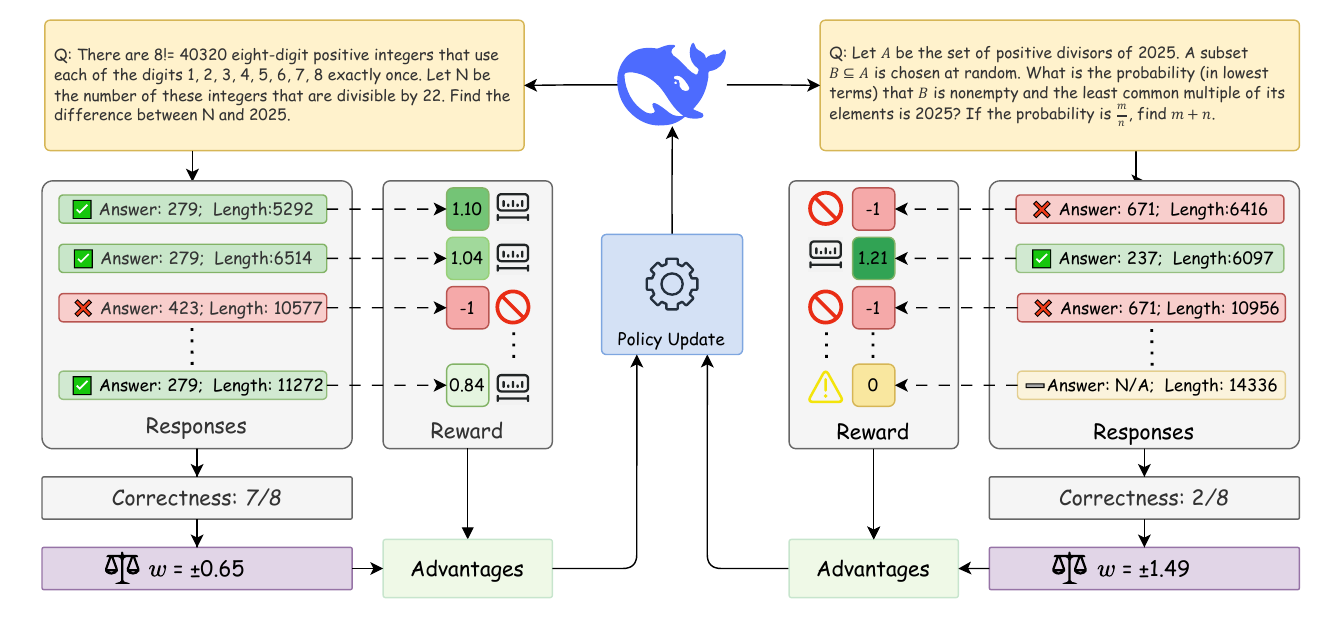} 

    \caption{The \textsc{GRPO-LEAD} framework assigns length-regularized positive rewards to correct answers and explicit penalties to incorrect ones. A difficulty-based weight $w$ used for advantage reweighting is determined from the empirical correctness of responses for each question. This weight then scales the advantages derived from each question, prioritizing harder questions over easier ones during the policy update to foster robust reasoning.}

     \label{fig:method} 

 \end{figure*}

% However, existing GRPO implementations still encounter substantial limitations. One key issue is the inherent reward sparsity arising from binary and rule-based accuracy metrics, which significantly hampers effective model training.  Specifically, when all generated responses to a given problem are either uniformly correct or incorrect, the resulting uniform reward signal provides minimal differentiation, leading to weak learning gradients and consequently slower model convergence. To elaborate, if all outputs in a question group are correct, each receives identical positive feedback, diluting the informative gradient needed for meaningful policy improvements. Conversely, uniformly incorrect responses yield no useful information to guide policy refinement. 

Recently, R1-like reasoning models have attracted significant attention due to their impressive performance in solving challenging mathematical reasoning tasks through extensive chains of thought~\cite{deepscaler2025,wen2025light}. According to the technical report introducing R1~\cite{guo2025deepseek}, reinforcement learning (RL) fine-tuning plays a pivotal role in enabling this reasoning capability. In particular, Group Relative Policy Optimization (GRPO)~\cite{shao2024deepseekmath}, a novel RL approach for language models, has emerged as a promising alternative to traditional methods such as PPO~\cite{schulman2017proximal} and DPO~\cite{rafailov2023direct}, primarily due to its efficiency and intrinsic compatibility with language model training. Researchers across various domains have successfully employed GRPO~\cite{grpo-1,grpo-2,grpo-3, grpo4}, achieving impressive outcomes.

Despite its strengths, existing GRPO implementations encounter significant limitations. A primary issue is reward sparsity stemming from binary, rule-based accuracy metrics; when responses within problem groups exhibit uniform correctness or incorrectness, the resulting uniform reward signals offer minimal differentiation, weakening learning gradients and hampering convergence. Moreover, such uniform signals inadequately promote concise reasoning, leading to unnecessarily verbose outputs and inefficiencies during training and inference. Additionally, the current reward formulation lacks explicit penalties for incorrect answers~\cite{incorrect-reward1,deepscaler2025,incorrect-reward2}, inadvertently encouraging models to guess rather than engage in rigorous reasoning, thereby compromising precision. Furthermore, rewards are applied uniformly across problems regardless of their intrinsic difficulty, causing models to excessively optimize simpler tasks while neglecting more challenging problems that require deeper reasoning.

Furthermore, computational efficiency also emerges as a critical practical concern, as reinforcement learning fine-tuning typically demands substantial resources, limiting accessibility, experimentation speed, and scalability, especially in low-resource environments. The current GRPO formulation is insufficient for encouraging concise and precise reasoning. Consequently, reducing computational requirements during both training and inference is essential for enabling broader applicability and effective real-world deployment.

Motivated by these limitations, this work introduces \textsc{GRPO-LEAD}, a suite of targeted modifications explicitly designed to enhance GRPO's effectiveness for mathematical reasoning tasks. The overall framework is illustrated in Figure \ref{fig:method}. Our key contributions include:
\begin{itemize}\setlength\itemsep{-0.2em}
\item We introduce a length-regularized reward with an explicit penalty for incorrect solutions to encourage solution conciseness while maintaining accuracy.
\item We apply difficulty-aware advantage reweighting to focus learning on more challenging problems, fostering robust generalization.
\item Our comprehensive evaluations demonstrate \textsc{GRPO-LEAD} significantly improves reasoning accuracy and conciseness, achieving state-of-the-art performance in mathematical reasoning for 14B-scale models.
\end{itemize}
% \begin{enumerate}
%     \item \textbf{Length-dependent Accuracy Reward:} Introduces a dynamic reward shaping mechanism that promotes brevity among correct responses using standardized length-based penalties, reducing verbosity without sacrificing accuracy.

%     \item \textbf{Explicit Penalty for Incorrect Solutions:} Implements a negative reward for incorrect outputs to enforce a sharper decision boundary, mitigating overconfidence and boosting precision of the model.

%     \item \textbf{Difficulty-aware Advantage Reweighting:} Applies a logistic weighting function to advantage estimates based on empirical correctness rates, ensuring stronger updates for harder problems and balanced generalization.

%     \item \textbf{Impact of Model Scale and Data Quality on Reinforcement Learning Effectiveness:} Demonstrates that larger base models and high-quality, curriculum-structured fine-tuning data significantly improve RL convergence and output quality; also introduces targeted reward interventions to mitigate repetitive formatting artifacts.
% \end{enumerate}

% Empirical evaluations and comprehensive ablation studies confirm that our method effectively addresses previous GRPO shortcomings, leading to more concise, accurate, and efficiently trainable models capable of robust performance across mathematical reasoning tasks.

\section{Related Work}
\subsection{ Group Relative Policy Optimization}
Group Relative Policy Optimization (GRPO) is a recently proposed algorithm designed specifically for fine-tuning language models with group-level normalization of rewards~\cite{guo2025deepseek}. GRPO modifies the standard policy gradient objective by introducing relative advantages within sets of responses corresponding to the same query, stabilizing updates and promoting consistent learning signals. Formally, GRPO defines the objective as:
\begin{align}
\mathcal{L}_{\text{GRPO}}(\theta) = 
\frac{1}{G} \sum_{i=1}^{G} \frac{1}{|o_i|} \sum_{t=1}^{|o_i|}
\Big[ 
\min \big( r_{i,t}(\theta) \hat{A}_{i,t},\, \\
\text{clip}(r_{i,t}(\theta), 1 - \epsilon, 1 + \epsilon) \hat{A}_{i,t} \big)
\Big] \nonumber
\label{eq:grpo}
\end{align}
where the importance sampling ratio is given by
\begin{equation}
r_{i,t}(\theta) = 
\frac{\pi_{\theta}(o_{i,t} \mid q, o_{i,<t})}
     {\pi_{\theta_{\text{old}}}(o_{i,t} \mid q, o_{i,<t})}.
\label{eq:ratio}
\end{equation}

Here, \( G \) denotes the number of groups (e.g., different queries), \( \hat{A}_{i,t} \) is the normalized advantage within group \( i \), and \( \epsilon \) defines the clipping range for conservative updates.

\subsection{Length Reward}
% A prevalent issue in reinforcement learning-based fine-tuning of language models is the uncontrolled increase in response length driven by reward signals, commonly known as \textit{reward hacking}\cite{reward-1,reward-2,reward-3}.  
A prevalent issue in reinforcement learning-based fine-tuning of language models is \textit{reward hacking}~\cite{reward-1,reward-2,reward-3}. In GRPO, when the model is trained with a large fixed budget, it can exploit this budget by producing an excessive number of extra reasoning and verification steps to ensure the correctness of the answer and therefore reach a higher reward. This phenomenon leads to unnecessarily verbose responses that lack conciseness and hinder interpretability, resulting in inefficiency in reasoning and reducing the model's practicality.
% Furthermore, such verbosity fails to reflect efficient reasoning, limiting model utility in practical scenarios. 

Existing efforts to mitigate this problem typically involve incentivizing shorter answers to encourage more succinct reasoning processes. For example, Kimi proposed an individual min-max normalized length reward based on the lengths of generated responses \cite{team2025kimi}. Yeo \textit{et al.} introduced a cosine length reward function with fixed maximum and minimum thresholds to manage response lengths \cite{yeo2025demystifying}. Aggarwal \textit{et al.} utilized a target "golden length" to directly reward or penalize responses based on their deviation from an ideal length~\cite{aggarwal2025l1}.

However, these existing methods depend heavily on static or predefined length heuristics, limiting their effectiveness across diverse questions of varying complexity. In contrast, our proposed length-dependent accuracy reward addresses these limitations by dynamically calibrating rewards according to each group’s relative response length and rollout accuracy, promoting concise yet difficulty-aware reasoning processes.

\section{Method}
To systematically address the limitations identified in existing implementations of Group Relative Policy Optimization (GRPO), we propose a suite of novel modifications collectively termed \textbf{ GRPO-LEAD} (\textbf{GRPO} with \textbf{L}ength-dependent rewards, \textbf{E}xplicit penalties, and \textbf{A}dvantage reweighting for \textbf{D}ifficulty). Our proposed method enhances the original GRPO framework by introducing three core innovations: 1) a length-dependent accuracy reward to foster concise solutions, 2) an explicit penalty mechanism to mitigate low precision rate caused by length reward, and 3) a difficulty-aware advantage reweighting strategy that amplifies learning signals for challenging problems. Additionally, we examine how base model scale and supervised fine-tuning (SFT) impact the effectiveness of reinforcement learning (RL) fine-tuning.

\subsection{Length-Dependent Accuracy Reward}

The core idea is to reward correct completions not uniformly but in proportion to their relative conciseness. Given a question $q$ and a set of model-generated responses $\{o_i\}$, we first isolate the subset of correct responses and compute the mean $\mu$ and standard deviation $\sigma$ of their token lengths. For a correct response $o$ with length $|o|$, we define its standardized length deviation as:
\begin{equation}
    z = \frac{|o| - \mu}{\sigma + \epsilon},
\end{equation}
where $\epsilon > 0$ is a small constant added for numerical stability. The final reward is modulated using an exponential decay function:
\begin{equation}
    R_{\text{accuracy}}(o|q) = 
    \begin{cases}
        \exp(-\alpha z), & \text{if } o \text{ is correct},\\
        0, & \text{if } o \text{ is incorrect}.
    \end{cases}
\end{equation}
where $\alpha > 0$ is a tunable hyperparameter controlling the strength of length penalization.

This formulation ensures that overly long correct responses are systematically penalized, while relatively concise ones are amplified. Unlike static or absolute length constraints, our approach leverages standardized deviation, allowing for dynamic adaptation to the distributional properties of each question.

% This reward function encourages models to optimize not only for correctness but also for brevity, enhancing alignment with human preferences for clear and efficient reasoning. In addition, this reward function also helps the model to converge faster as it is less sparse than original accuracy reward. Empirically, we observe that the method reduces verbosity without degrading task performance too much, particularly in settings where concise responses are preferable.

\vspace{1mm}
\subsection{Explicit Penalty for Incorrect Answers to Enhance True Accuracy}

% Many existing methods prioritize maximizing \textit{pass@1}—the success rate on the first attempt—often within constrained response lengths. However, our observations reveal a critical trade-off: an overemphasis on \textit{pass@1} can inadvertently degrade overall model accuracy. While length-based regularization, which might curtail the model's capacity for extended reasoning or self-correction, contributes to this, the fundamental issue appears to stem from the use of a binary accuracy reward.

% Disabling length-based rewards can improve \textit{pass@1} during both training and evaluation. Yet, this alone does not resolve the underlying problem. A more holistic metric, such as \textit{solve-all}—the proportion of questions for which \textit{all} sampled responses are correct—often declines. The reintroduction of a length reward merely accelerates this negative trend. We attribute this phenomenon to the incentive structure of binary rewards: under pressure to generate responses within a limited length, a model is encouraged to provide an answer, even if it's a guess, rather than no answer at all. While such behavior might yield a non-zero reward for ambiguous or partially complete outputs, thereby inflating \textit{pass@1}, it ultimately diminishes overall precision by rewarding less rigorous reasoning.
Existing methods often prioritize maximizing \textit{pass@1}—the success rate on the first attempt—typically within restricted response lengths. However, this focus can inadvertently degrade overall model accuracy. The fundamental issue appears to stem from the use of a binary accuracy reward, rather than length-based regularization: under pressure to generate responses within a limited length, a model is encouraged to provide an answer, even if it's a guess, rather than no answer at all. Such guesses can achieve a non-zero reward and inflate \textit{pass@1}, but they do so at the cost of overall precision by rewarding less rigorous reasoning. 

To counteract this tendency and foster a more robust distinction between correct and incorrect outputs, we introduce a revised reward structure that explicitly penalizes incorrect responses. This new reward function is defined as:
\begin{equation}
    R_{\text{accuracy}}(o \mid q) =
    \begin{cases}
        \exp(-\alpha z), & \text{if } o \text{ is correct}, \\
        -1, & \text{if } o \text{ is incorrect},
    \end{cases}
\end{equation}
where $o$ is the output, $q$ is the question, $z$ represents the standardized length deviation of a correct response, and $\alpha > 0$ is a hyperparameter controlling the strength of the length penalization for correct answers, consistent with prior definitions.

The expected reward for a response, given its probability of correctness $P(\text{correct})$, under this formulation is:
\begin{align}
    \mathbb{E}[R_{\text{accuracy}}(o \mid q)]
    &= P(\text{correct}) \cdot \exp(-\alpha z) \notag \\
    &\quad - (1 - P(\text{correct})) \tag{4}
\end{align}
To intuitively grasp the impact of this reward function, let us consider a simplified scenario where the length penalty for correct answers is negligible (i.e., $\exp(-\alpha z) \approx 1$). In practice, the average reward for correct answers often normalizes close to this value. Under this assumption, the expected reward simplifies to:
\begin{equation}
    \mathbb{E}[R] \approx 2P(\text{correct}) - 1
    \tag{5}
\end{equation}
This approximation reveals a crucial characteristic: the expected reward becomes positive only when $P(\text{correct}) > 0.5$. This threshold acts as a principled deterrent against speculative guessing, compelling the model to internalize a more stringent decision boundary for correctness. Our empirical results confirm that this approach significantly improves both \textit{pass@1} and overall precision, encouraging the model to favor accuracy over mere completion.

\subsection{Advantage Reweighting for Difficulty-Aware Training}

While length reward and advantage reweighting can enhance precision and mitigate verbosity, uniformly applying rewards across all questions, irrespective of their intrinsic difficulty, may implicitly bias the model. It might learn to excessively optimize performance on simpler tasks-—where correct and concise responses are more readily achieved-—while neglecting more complex questions that demand deeper reasoning. Consequently, the performance on challenging problems can degrade.

Therefore, we introduce a difficulty-aware advantage reweighting strategy, which dynamically adjusts the magnitude of policy updates based on an estimate of problem difficulty. The intuition is to amplify learning signals for harder tasks, re-anchoring the model towards harder tasks.

Formally, we first quantify problem difficulty. For a given question $q$ and its associated set of sampled responses $\{o_i\}$, we define the group's empirical correctness ratio as:
\begin{equation}
\rho_q = \frac{\text{number of correct responses for } q}{\text{total number of responses for } q}.
\end{equation}
This ratio, $\rho_q$, serves as an inverse proxy for problem difficulty: a lower $\rho_q$ suggests a harder question.

Next, we introduce a logistic reweighting factor dependent on this ratio to modulate the advantage estimates during the RL training step. The logistic function is defined as:
\begin{equation}
w(\rho_q) = A + \frac{B - A}{1 + \exp\left[k(\rho_q - \rho_0)\right]},
\end{equation}
where hyperparameters $A, B, \rho_0, k$ allow precise control over the sensitivity of weighting to problem difficulty.

To apply this reweighting, we first consider the normalized advantage estimate for a response $o_i$ to question $q$:
\begin{equation}
\tilde{A}_i = \frac{R(o_i|q) - \mu_q}{\sigma_q + \epsilon},
\end{equation}
where $\mu_q$ and $\sigma_q$ are the mean and standard deviation of rewards $R(o_i|q)$ for responses to question $q$, and $\epsilon$ is a small constant for numerical stability. We then define the difficulty-aware advantage, $A_i^\prime$, as:
\begin{equation}
A_i^\prime = \tilde{A}_i \cdot
\begin{cases}
w(\rho_q), & \text{if } \tilde{A}_i > 0 \\
w(1 - \rho_q), & \text{if } \tilde{A}_i \leq 0
\end{cases}
\end{equation}
This formulation ensures that for difficult problems (low $\rho_q$), correct responses (which are rare and thus highly valuable) receive substantially larger updates due to the increased weighting $w(\rho_q)$. Conversely, incorrect responses on easier problems (high $\rho_q$) are penalized more strongly, sharpening the decision boundary for problems where high performance should be expected.

% Empirically, the difficulty-aware reweighting leads to more balanced model performance across tasks of varying complexity. This advantage reweighting method complements the previously introduced reward strategies, collectively promoting concise, accurate, and difficulty-adaptive reasoning in language models.

\subsection{Impact of Data Quality on Reinforcement Learning Effectiveness}
To further enhance model capabilities, we first performed supervised fine-tuning (SFT) on a specialized dataset of 13k math reasoning problems sourced from DeepScaler~\cite{deepscaler2025} (including historical AMC, AIME, and OmniMath problems) with solutions generated by QwQ32B~\cite{qwq32b}. Although this SFT model initially showed signs of overfitting, subsequent application of our proposed RL strategies rapidly mitigated these issues. This SFT+RL approach yielded faster convergence and significantly improved pass@1 accuracy and overall precision compared to applying RL directly to the original base model.

Our findings also highlight the critical role of data quality and curriculum strategies in RL. We established a robust initial policy by applying RL to a subset of challenging problems from the DeepScaler dataset. This policy was then further refined using a curriculum composed of the most challenging problems identified from this first RL stage and supplemented by high-difficulty examples from the Light-R1 dataset~\cite{wen2025light}. This two-stage curriculum markedly enhanced the model's ability to continuously improve on complex tasks.

Finally, we addressed a persistent formatting issue of repetitive n-gram patterns, likely stemming from an absence of clear end-of-sequence (EOS) signals during SFT. By temporarily removing length-dependent rewards and introducing an explicit negative reward ($-1.5$) for such repeated n-grams, we achieved further improvements in precision and pass@1 metrics. This intervention demonstrates the effectiveness of targeted reward modifications for mitigating specific output anomalies.

In summary, our experiments affirm that initial model capacity, curated data curricula for RL, and targeted reward engineering are pivotal for optimizing fine-tuning outcomes. These elements collectively inform a systematic approach for enhancing language models' ability to produce concise, accurate, and well-structured responses across tasks of varying complexity.

\section{Experimental Setup}

We evaluate \textsc{GRPO-LEAD}, integrating length-dependent accuracy rewards, explicit penalties for incorrect solutions, and difficulty-aware advantage reweighting, on \textsc{DeepSeek-R1 Distilled} variants~\cite{guo2025deepseek,yang2024qwen25}. Our experiments cover two model scales, 7B and 14B parameters. All GRPO training is conducted using the VERL framework.\cite{sheng2024hybridflow}.

\subsection{Datasets and Filtering}

Our primary training data is sourced from the \textsc{DeepScaler} dataset~\cite{deepscaler2025}. We filter out problems with difficulty ratings below 2.5, resulting in approximately 9{,}000 questions for fine-tuning.

For stage 2 of our 14B model experiments, we further refine the dataset by selecting problems where the model's stage-1 rollout accuracy is no greater than 75\%, yielding around 2{,}283 questions. Additionally, we incorporate challenging problems with numeric answers from the stage-2 dataset of Light-R1~\cite{wen2025light}, resulting in 3{,}524 questions in total.

% In total, the dataset for stages 2 comprises 3{,}524 questions. This adaptive filtering strategy ensures a focused emphasis on harder problems, aiming to improve the model’s performance on more complex tasks.

\subsection{Hyperparameters}

We train with a learning rate of \(1 \times 10^{-6}\), batch size 32, and group size 8---generating 8 rollouts per question for GRPO reward computation. The KL penalty term is removed, as it was found to suppress exploration in our experiments, which is also suggested in similar works~\cite{liu2025understanding, hu2025open}. 

For the length-dependent accuracy reward, we set \(\alpha = 0.05\), providing a moderate decay that encourages conciseness without penalizing slight verbosity. For difficulty-aware advantage reweighting, we use \(A = 0.4\), \(B = 1.5\), \(\rho_0 = 0.75\), and \(k = 10\). This configuration ensures reweighting is minimal on easy problems but increases sharply near the 75\% correctness threshold. The steep slope (\(k = 10\)) enables strong emphasis on high-difficulty examples, guiding the model to allocate learning more effectively.

\subsection{Model Variants and Fine-Tuning Stages}

\paragraph{7B Model Experiments}
Starting from the DeepSeek-R1 Distilled 7B Qwen-Math checkpoint, we first apply standard GRPO on the 9k questions, producing a baseline. Then, we train three more models from the DeepSeek-R1 Distilled 7B Qwen-Math checkpoint, adding one more of the following components subsequently:
(i) Length Reward only,
(ii) Length Reward + Advantage Reweighting,
(iii) Length Reward + Advantage Reweighting + Explicit Penalty.
We train for approximately 200 steps and select the top-performing checkpoints based on validation results. At test time, we limit the generation length to 8k for all 7B models, matching the training length limit. 

\paragraph{14B Model Experiments}
We extend the above procedure to the DeepSeek-R1 Distilled 14B Qwen checkpoint across multiple stages. In \textbf{Stage~1}, we train for 100 steps using all \textsc{GRPO-LEAD} components on the filtered 9k-question dataset. To enhance the model's base capability, we first fine-tune the model on a curated set of 13k math problems with supervised fine-tuning (SFT), then conduct the RL phase. This SFT stage significantly improves the model’s reasoning quality, even though it tends to increase the output length and caused some format errors.

The SFT data consists of all problems in the \textsc{DeepScaler} dataset with difficulty greater than 1. To construct high-quality reasoning traces for SFT, we use the QWQ-32B model~\cite{qwq32b} to generate step-by-step solutions.

After observing that some questions remain low correctness, we further fine-tune for \textbf{Stage~2} to focus on those underperformed problems. We also address the repetitive output patterns by removing the length penalty and introducing a negative reward ($-1.5$) for repeated $n$-grams. We continue training for 240 more steps (100 steps with initial settings and 140 more steps with repetition penalty), yielding the final model checkpoint. At test time, we limit the generation length to 14k for all 14B models, in accordance with our training settings and also to better compare the models' performance in a low-budget scenario.

\subsection{Baselines and Evaluation Protocol}
We compare our models with both \textsc{DeepSeek-R1 Distilled-14B-Qwen}~\cite{guo2025deepseek} (the distilled Qwen model without \textsc{GRPO-LEAD}) and \textsc{Light-R1-14B-DS}~\cite{wen2025light}, which has the same base model as ours and was first finetuned with 3k hard math problems with SFT, and then fine-tuned with a cosine-based length reward~\cite{yeo2025demystifying} on their selected math problems for three epochs using GRPO.

We primarily report three metrics:
\textbf{(1) Cons@32}, accuracy through majority voting for 32 samplings;
\textbf{(2) Pass@1}, the probability that the top-1 sample is correct under a chosen decoding strategy;
\textbf{(3) Average Length} (Len$_{\mathrm{avg}}$), measuring verbosity.
Unless otherwise specified, we decode with temperature 0.6 and sample 32 solutions per question, then compute Cons@32 and Pass@1 over these samples.

\section{Results}
\label{sec:results}

\begin{table*}[h]
\centering
\small
\renewcommand{\arraystretch}{1.05}
\setlength\tabcolsep{6pt}

\begin{tabular}{lccc ccc}
\toprule
\textbf{Ablation Setting} 
& \multicolumn{3}{c}{\textsc{AIME24}} 
& \multicolumn{3}{c}{\textsc{AIME25}} \\
\cmidrule(lr){2-4} \cmidrule(lr){5-7}
& \textbf{Cons@32} & \textbf{Pass@1} & \textbf{Len$_{\mathrm{avg}}$}
& \textbf{Cons@32} & \textbf{Pass@1} & \textbf{Len$_{\mathrm{avg}}$} \\
\midrule
Deepseek-7B
& \underline{0.767} & 0.431 & 6,990 
& 0.467 & 0.292 & 7,113 \\
\midrule
GRPO + len. reward 
& \underline{0.767} & 0.438 & \textbf{5,275} & 0.533 & 0.308 & \textbf{5,210} \\
\quad \quad \quad + adv. reweighting
& \underline{0.767} & \underline{0.458} & \underline{5,323} & \textbf{0.567} & \underline{0.325} & \underline{5,437} \\
\quad \quad \quad+ explicit penalty 
& \textbf{0.800} & \textbf{0.470} & 6,104 & \textbf{0.567} & \textbf{0.345} & 6,308 \\
\bottomrule
\end{tabular}
\caption{Ablation results on \textsc{AIME24} and \textsc{AIME25}. We report \textbf{Cons@32} (accuracy through majority voting for 32 samplings), \textbf{Pass@1}, and the average token length (\textbf{Len$_{\mathrm{avg}}$}). The best value in each column is in boldface, the second best is underlined.}
\label{tab:merged_model_performance_tasks}
\end{table*}

In this section, we present a comprehensive evaluation of the proposed \textsc{GRPO-LEAD} framework on two mathematical benchmarks: \textsc{AIME24} and \textsc{AIME25}. Our analysis is structured as follows: we first examine training dynamics to illustrate how \textsc{GRPO-LEAD} accelerates convergence; next, we perform an ablation study to assess the incremental benefits of each component; and finally, we compare against state-of-the-art baselines for 14B-scale language models.

\subsection{Training Dynamics}

\begin{figure}
     \centering
    \includegraphics[width=1\linewidth]{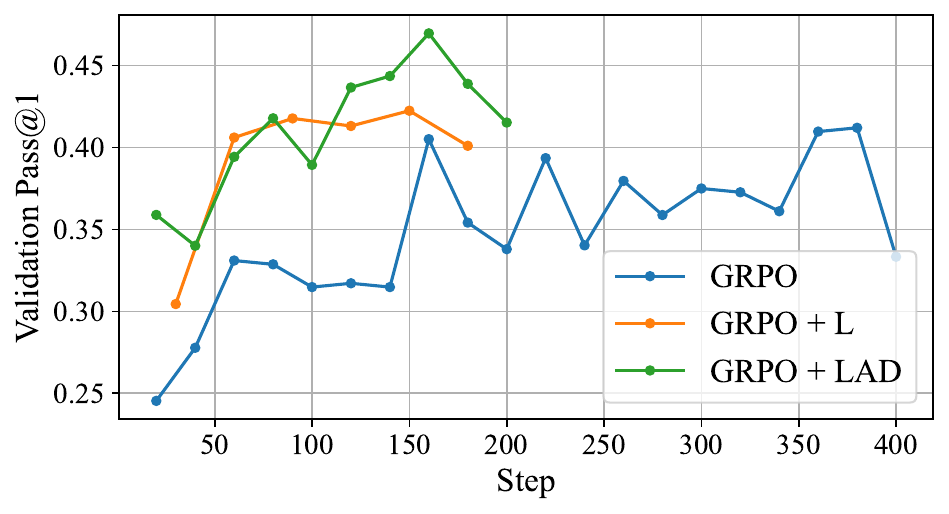}
    \caption{Validation$^*$ Pass@1 over training steps for three configurations: GRPO, GRPO+L, and GRPO+LAD. As shown by the faster convergence, length reward and advantage reweighting provide a richer reward signal signal than the original setup.}
    \label{fig:Validation}
\end{figure}

\footnotetext[1]{The validation consists of 27 challenging problems from AIMO2~\cite{ai-mathematical-olympiad-progress-prize-2}, CMU-MATH-AIMO~\cite{Sun}, and AIME24.} 

Figure~\ref{fig:Validation} plots the evolution of Pass@1 on a validation split over training steps for three configurations of the 7B model: (i) baseline GRPO, (ii) GRPO with length reward, and (iii) GRPO with both length reward and advantage reweighting. We observe two clear trends. First, adding a length-dependent reward not only yields higher Pass@1 but also accelerates early-stage convergence, suggesting that penalizing overly verbose correct solutions provides a more informative learning signal. Second, incorporating advantage reweighting (to amplify updates on harder questions) further steepens the trajectory, indicating that reweighting advantage estimates according to problem difficulty helps the model refine reasoning on challenging prompts more efficiently.

Overall, these dynamics confirm that \textsc{GRPO-LEAD} components---particularly the length reward---bolster training stability and speed. By comparison, the baseline GRPO model learns more slowly and lags behind in Pass@1 across the entire training horizon.

\subsection{Ablation Analysis}

\begin{figure*}[ht]
    \centering
    
    \begin{subfigure}[b]{0.45\textwidth}
        \centering
        \includegraphics[width=\textwidth]{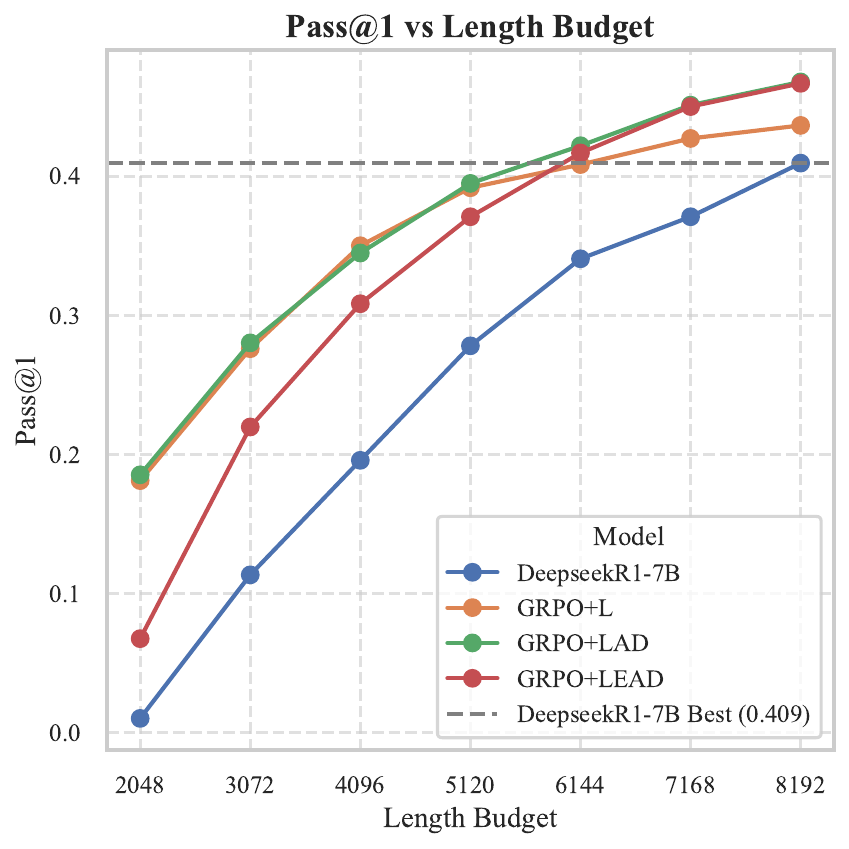}
        \caption{AIME24}
        \label{fig:sub1}
    \end{subfigure}
    \hfill
    \begin{subfigure}[b]{0.45\textwidth}
        \centering
        \includegraphics[width=\textwidth]{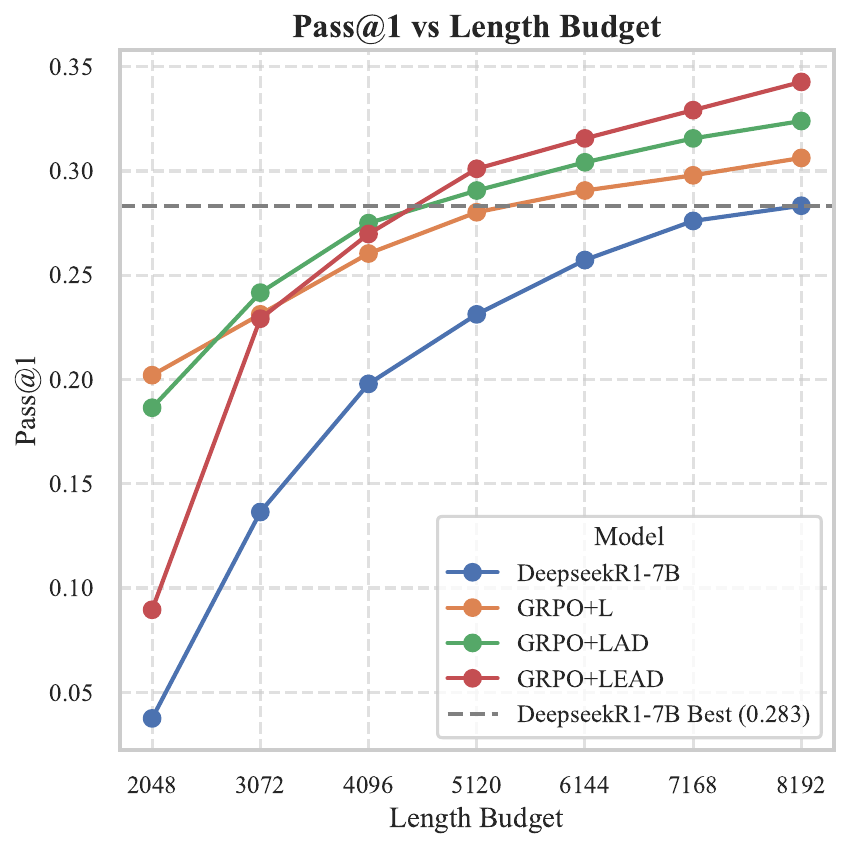}
        \caption{AIME25}
        \label{fig:sub2}
    \end{subfigure}
    
    \caption{Performance against inference budget for training done with different ablations of LEAD. GRPO with length reward (GRPO+L) largely enhances the performance at low budget settings compared to before training (DeepseekR1-7B). }
    \label{fig:budget}
\end{figure*}

We next quantify the contribution of each \textsc{GRPO-LEAD} component through a step-by-step ablation on the 7B model. Table~\ref{tab:merged_model_performance_tasks} summarizes results on \textsc{AIME24} and \textsc{AIME25}.

\paragraph{Length Reward Brings Conciseness to Reasoning}
We first incorporate the length-dependent accuracy reward into GRPO. Compared to Deepseek-7B, length reward slightly improves Pass@1 on both \textsc{AIME24} by 1.6\% ($0.431\rightarrow0.438$)  and \textsc{AIME25} by 5.4\% ($0.292\rightarrow0.308$), with an additional improvement of Cons@32 by 14.1\% on \textsc{AIME25}. Notably, these improvements are accompanied by a substantial reduction of 1,715 tokens (24.5\%) and 1,903 tokens (26.8\%) in the average response length on the two datasets, respectively. Figure~\ref{fig:budget} further demonstrates that length reward largely enhances performance in low-budget settings over the base model, matching its peak performance with only 5/8 of the token budget on the more difficult \textsc{AIME25}. These results demonstrate that length reward, by penalizing correct but overly verbose solutions, can effectively reduce unnecessary text without compromising overall performance.

\paragraph{Advantage Reweighting Encourages Model to Solve More Difficult Problems}
Further incorporating difficulty-aware advantage reweighting (GRPO+LAD) refines performance. On \textsc{AIME24}, Pass@1 increases from the GRPO+L stage by 4.8\% ($0.438\rightarrow0.458$), while Cons@32 remains $0.767$. For \textsc{AIME25}, both Pass@1 and Cons@32 improve by 5.5\% ($0.308\rightarrow0.325$) and 6.4\% ($0.533\rightarrow0.567$), respectively. As Figure~\ref{fig:budget} shows, GRPO+LAD demonstrates gains over GRPO+L in almost all budget regimes on \textsc{AIME25} and for budgets exceeding 5k tokens on \textsc{AIME24}. These results indicate that advantage reweighting, by prioritizing challenging problems, strengthens reasoning robustness and mitigates over-reliance on simpler examples, thus validating its role in driving more reliable generalization.

\begin{table*}[h]
\centering
\small
\renewcommand{\arraystretch}{1.05}
\setlength\tabcolsep{6pt}

\begin{tabular}{lccc ccc}
\toprule
\textbf{Model Name} 
& \multicolumn{3}{c}{\textsc{AIME24}} 
& \multicolumn{3}{c}{\textsc{AIME25}} \\
\cmidrule(lr){2-4} \cmidrule(lr){5-7}
& \textbf{Cons@32} & \textbf{Pass@1} & \textbf{Len$_{\mathrm{avg}}$}
& \textbf{Cons@32} & \textbf{Pass@1} & \textbf{Len$_{\mathrm{avg}}$} \\
\midrule
DeepSeek-14B 
& 0.800 & 0.614 & 9,182 
& 0.633 & 0.429 & 10,046 \\
Light-R1-14B-DS 
& \underline{0.833} & \underline{0.641} & 9,571 
& \textbf{0.767} & 0.505 & 10,194 \\
LEAD-stage1 
& \underline{0.833} & 0.629 & \underline{8,790} 
& \textbf{0.767} & \underline{0.523} & \underline{9,371} \\
LEAD-stage2
& \textbf{0.867} & \textbf{0.650} & \textbf{8,267} 
& \textbf{0.767} & \textbf{0.539} & \textbf{8,668} \\
\bottomrule
\end{tabular}

\caption{Comparison of model performance on \textsc{AIME24} and \textsc{AIME25}, showing \textbf{Cons@32}, \textbf{Pass@1}, and average token length (\textbf{Len$_{\mathrm{avg}}$}). The best value in each column is in boldface, the second best is underlined.}
\label{tab:merged_model_metrics_grouped}
\end{table*}

\paragraph{Explicit Penalty for Incorrect Answers Regularizes Thinking}
Finally, introducing an explicit penalty for incorrect solutions (GRPO+LEAD) yields the highest Pass@1 scores. On \textsc{AIME24}, Pass@1 and Cons@32 improve from the GRPO+LAD stage by 2.6\% ($0.458\rightarrow0.470$) and 4.3\% ($0.767\rightarrow0.800$), respectively. On \textsc{AIME25}, Pass@1 also increases by 6.2\% ($0.325\rightarrow0.345$), as detailed in Table~\ref{tab:merged_model_performance_tasks}. Notably, these gains involve a modest increase in average solution length on \textsc{AIME24} (from approximately 5,300 to 6,104 tokens). Figure~\ref{fig:budget} illustrates this trade-off, showing a performance sacrifice in low-budget regimes, though GRPO+LEAD still outperforms GRPO+LAD with budgets higher than 5k tokens on \textsc{AIME25}. These results suggest that the explicit penalty serves as a regularizer for the model to be more conservative about its reasoning. Such regularization boosts performance while requiring a slightly longer thinking process, which nevertheless remains shorter than the Deepseek-7B baseline.

\paragraph{}
Overall, these ablation results confirm that all three enhancements---length-dependent accuracy, difficulty-aware advantage reweighting, and explicit penalties---collectively reduce verbosity, strengthen mathematical skills on harder questions, and elevate precision in final predictions.

\subsection{Comparison with Baselines}

We next evaluate \textsc{GRPO-LEAD} at the 14B scale and compare it against two strong baselines under a 14k-token generation budget: \textbf{DeepSeek-14B} and the state-of-the-art \textbf{Light-R1-14B-DS}. Table~\ref{tab:merged_model_metrics_grouped} presents results on \textsc{AIME24} and \textsc{AIME25}, including both our intermediate model (\textit{LEAD-stage1}) and our final model (\textit{LEAD-stage2}).

\paragraph{AIME24 Performance} LEAD-stage1 achieves a Cons@32 of 0.833, matching Light-R1-14B-DS and exceeding DeepSeek-14B by 4.1\%. Its Pass@1 outperforms DeepSeek-14B by 2.4\% and closely approaches Light-R1-14B-DS. Crucially, LEAD-stage1 produces more concise responses than both baselines, with more than 800 tokens less on average. Building on these gains, LEAD-stage2 pushes performance further, delivering the highest Cons@32 (4\% above Light-R1-14B-DS) and the best Pass@1, while reducing average solution length to 8,267 tokens.

\paragraph{AIME25 Performance} LEAD-stage1 yields a Cons@32 of 0.767, matching Light-R1-14B-DS and exceeding DeepSeek-14B by 21.2\%. Its Pass@1 (0.523) outperforms DeepSeek-14B by 21.9\% and Light-R1-14B-DS by 3.6\%. Crucially, LEAD-stage1 produces more concise responses than both baselines, with its solutions averaging 9,371 tokens. Building on these gains, LEAD-stage2 pushes performance further, delivering the highest Cons@32 (matching Light-R1-14B-DS at 0.767) and the best Pass@1 (0.539), while reducing average solution length to 8,668 tokens.

Overall, both LEAD-stage1 and LEAD-stage2 deliver substantial improvements over DeepSeek-14B and Light-R1-14B-DS, simultaneously boosting correctness and conciseness under a constrained (14k-token) budget. Remarkably, training LEAD-stage1 for just 100 steps—requiring only about 24 hours on eight H20 GPUs—already matches Light-R1-14B-DS on Cons@32 and outperforms it on AIME25 Pass@1 while producing shorter solutions, underscoring the practical efficiency of \textsc{GRPO-LEAD} for large-scale math problem-solving.

\section{Conclusion}
\label{sec:conclusion}

% We have introduced \textsc{GRPO-LEAD}, an enhanced reinforcement learning framework tailored for mathematical reasoning tasks. Built upon Group Relative Policy Optimization, \textsc{GRPO-LEAD} augments the reward structure with three key components: a length-dependent accuracy reward to discourage verbosity, an explicit negative penalty that enforces stricter decision boundaries between correct and incorrect solutions, and a difficulty-aware advantage reweighting strategy to increase focus on challenging problems. Empirical results on two AIME benchmarks demonstrate that \textsc{GRPO-LEAD} not only accelerates convergence but also improves the model's reasoning ability while maintaining succinct solution paths. Our 14B-scale experiments also highlight that \textsc{GRPO-LEAD} outperforms state-of-the-art baselines by balancing concise output generation with high problem-solving accuracy. While open challenges remain---particularly in handling partial correctness and scaling to broader domains---our findings suggest that reward shaping and difficulty modeling are promising directions for making language models more aligned and robust in mathematical reasoning.
We introduced \textsc{GRPO-LEAD}, a reinforcement learning framework designed for mathematical reasoning tasks. By extending Group Relative Policy Optimization with three major components—(1) a length-dependent accuracy reward to discourage overly verbose solutions, (2) an explicit negative penalty that clarifies the boundary between correct and incorrect answers, and (3) a difficulty-aware advantage reweighting scheme to prioritize tougher problems—\textsc{GRPO-LEAD} addresses key challenges in structured problem-solving.

Empirical evaluations on two AIME benchmarks show that \textsc{GRPO-LEAD} not only speeds up convergence but also strengthens the model’s reasoning capability while keeping solution paths concise. Our 14B-scale experiments further confirm that \textsc{GRPO-LEAD} achieves state-of-the-art performance by balancing output brevity with high problem-solving accuracy. Although open questions remain—particularly in managing partial correctness and extending these techniques to broader domains—our findings suggest that reward shaping and difficulty modeling are pivotal in developing more robust and aligned language models for complex mathematical reasoning.
% We introduced \textsc{GRPO-LEAD}, a reinforcement learning framework tailored for mathematical reasoning, enhancing Group Relative Policy Optimization (GRPO) with three key innovations: (1) a length-dependent accuracy reward that discourages overly verbose solutions, (2) an explicit penalty clarifying the boundary between correct and incorrect answers, and (3) a difficulty-aware advantage reweighting strategy prioritizing challenging problems.

% Empirical evaluations on two AIME benchmarks demonstrate that \textsc{GRPO-LEAD} significantly accelerates convergence, improves reasoning capabilities, and ensures concise solution paths. Experiments at the 14B scale confirm \textsc{GRPO-LEAD} achieves state-of-the-art performance, effectively balancing brevity with high accuracy. While challenges remain—especially around managing partial correctness and broader domain applicability—our findings underscore the importance of reward shaping and difficulty modeling in developing robust, aligned models for complex mathematical reasoning.

\clearpage
\section{Limitations}
\label{sec:limitations}
Although our techniques for encouraging concise solutions and difficulty-balanced learning may transfer to other domains, the gains reported here are specific to mathematical reasoning tasks. Further studies are needed to evaluate the effectiveness of \textsc{GRPO-LEAD} on broader question-answering or logical reasoning domains, where correctness signals and domain structures can differ substantially. 

Additionally, we only have access to a limited amount of compute, which prevents us from conducting more comprehensive experiments. For instance, we currently cannot provide the validation curve for the 7B model in the ablation study that adds an explicit penalty. This is due to an error in the validation code after upgrading to the newest VERL version, and we currently do not have the compute to reproduce it. A comparison with the original GRPO model is also missing, except for the curve shown in Figure~\ref{fig:Validation}, because the checkpoint was stored on a rented server that was automatically released as we were writing the paper. We also couldn't formally perform a hyperparameter search to showcase the rationale behind choosing the hyperparameters for our designed modifications.

\newpage
\clearpage
% \bibliography{custom}
% \bibliographystyle{acl_natbib}

\newpage
\clearpage

\appendix

\begin{table*}[t] \centering \begin{tabular}{lccccc} \toprule \textbf{Model} & \textbf{Accuracy} & \textbf{Avg. Tokens (Overall)} & \textbf{Easy} & \textbf{Medium} & \textbf{Hard} \\ \midrule LEAD-14B & 0.5156 & 6322 & 3998 & 6912 & 8000 \\ DeepSeek-R1-Distill-Qwen-14B & 0.5103 & 5794 & 3046 & 6429 & 7856 \\ \bottomrule \end{tabular} \caption{Performance on LiveCodeBench (\texttt{release\_v5}) with maximum sequence length of 8k tokens. All token counts are rounded to the nearest integer.} \label{tab:code-results} \end{table*}

\section{Evaluations on Coding Tasks}

We evaluate our proposed \textbf{LEAD-14B} model against the original \textbf{DeepSeek-R1-Distill-Qwen-14B} baseline on the \textbf{LiveCodeBench} benchmark under a maximum sequence length of 8k tokens. The dataset version used is \texttt{release\_v5}, consisting of 880 code generation tasks. Results are summarized in Table~\ref{tab:code-results}.

\begin{table*}[h]
\centering

\begin{tabular}{lccccc}
\toprule
\textbf{Model} & \textbf{Cons@32} & \textbf{Avg. Correct} & \textbf{Avg. Answer} & \textbf{Precision} & \textbf{Pass@1} \\
\midrule
\multicolumn{6}{l}{\textit{Normal Problems (1--5)}} \\
\midrule
Deepseek-7B & 0.8 & 18.8 & 20.3 & 0.708 & 0.588 \\
GRPO + L & 0.8 & 19.7 & 27.6 & 0.631 & 0.616 \\
GRPO + LAD & \textbf{0.9} & 20.1 & 26.9 & 0.687 & 0.628 \\
GRPO + LEAD & 0.8 & 22.0 & 24.5 & \textbf{0.723} & \textbf{0.688} \\
\midrule
\multicolumn{6}{l}{\textit{Difficult Problems (6--10)}} \\
\midrule
Deepseek-7B & 0.4 & 8.3 & 13.8 & 0.404 & 0.259 \\
GRPO + L & 0.5 & 8.6 & 24.1 & 0.412 & 0.269 \\
GRPO + LAD & \textbf{0.6} & 9.8 & 24.2 & \textbf{0.448} & \textbf{0.306} \\
GRPO + LEAD & \textbf{0.6} & 9.7 & 20.0 & 0.421 & 0.303 \\
\midrule
\multicolumn{6}{l}{\textit{Highly Difficult Problems (11--15)}} \\
\midrule
Deepseek-7B & 0.2 & 0.9 & 2.0 & 0.230 & 0.028 \\
GRPO + L & 0.3 & 1.3 & 13.9 & 0.172 & 0.041 \\
GRPO + LAD & 0.2 & 1.3 & 14.6 & 0.156 & 0.041 \\
GRPO + LEAD & \textbf{0.4} & 1.5 & 7.7 & \textbf{0.355} & \textbf{0.047} \\
\bottomrule
\end{tabular}
\caption{Consolidated evaluation results on AIME25, stratified by problem difficulty. Avg. Answers refers to the number of outputs that have completed within the 8k token budget and produce some, where as Avg. Correct refers to the correct answers. Precision is Avg. Correct/ Avg. Answers.  }
\label{tab:aime_difficulty_combined}
\end{table*}

As shown above, \textbf{LEAD-14B} achieves higher accuracy (0.5156 vs.\ 0.5103) while producing slightly longer completions. This suggests that our method enhances reasoning capability in code generation. Regarding the observed increase in chain-of-thought (CoT) length, we hypothesize that this effect arises because our training focused exclusively on mathematical reasoning datasets. While our method compresses reasoning paths in math domains, such compression does not appear to generalize as effectively to code. Combined with the improved reasoning capability that may increase the overall reasoning path, this may explain why generated sequences are overall longer in code-related tasks.

\section{Detailed Analysis on AIME25 by Difficulty}
\label{sec:appendix_aime_difficulty}

To further analyze model performance, we stratified the AIME25 dataset into three difficulty levels based on the problem number: normal (problems 1--5), difficult (problems 6--10), and highly difficult (problems 11--15). The detailed evaluation results for each stratum are presented in Table~\ref{tab:aime_difficulty_combined}.

The stratified results in Table~\ref{tab:aime_difficulty_combined} support the hypothesis that advantage reweighting enhances a model's ability to solve more difficult problems. This is evidenced by the widening performance gap in Pass@1 between GRPO+L and GRPO+LAD as problem difficulty increases. For normal problems, GRPO+LAD offers a modest $1.95\%$ improvement over GRPO+L. This margin increases substantially to $13.7\%$ for difficult problems, indicating that the benefits of advantage reweighting are more pronounced in challenging scenarios.

For highly difficult problems, the Pass@1 scores for GRPO+L and GRPO+LAD are identical. Neither method incorporates an explicit penalty for incorrect answers, making them susceptible to generating numerous wrong solutions. This tendency leads to unstable majority voting-based accuracy (Cons@32), a vulnerability that is magnified by the intrinsic difficulty of the problems.

In contrast, the introduction of an explicit penalty in GRPO+LEAD demonstrates a clear regularization effect. On the most difficult problem set, GRPO+LEAD achieves the highest accuracy (Cons@32 of $0.4$) and more than doubles the precision of both GRPO+L ($0.172$) and GRPO+LAD ($0.156$); the number of correct answers generated by GRPO+LEAD is comparable to both GRPO+L and GRPO+LAD, despite generating much fewer total answers. This validates our hypothesis that the explicit penalty effectively "regularizes thinking", discouraging the kind of hasty and incorrect responses that the length reward tends to encourage otherwise.

\begin{table*}[!ht]
\centering
\small
\renewcommand{\arraystretch}{1.2}
\setlength\tabcolsep{6pt}
\begin{tabular}{p{0.2\linewidth} p{0.35\linewidth} p{0.35\linewidth}}
\toprule
\textbf{Aspect} & \textbf{GRPO+L} & \textbf{Deepseek-7B} \\
\midrule
\textbf{Structure \& flow} &
“\textbf{Step 1:} enumerate all possible triples … \textbf{Step 2:} compute the multinomial coefficient … \textbf{Step 3:} sum and mod.” &
“Okay … \textit{let me parse this step by step} … \textit{but wait, hold on} … \textit{let me verify the triples again} …” \\
\addlinespace
\textbf{Redundancy} &
“...hence, all possible triples: (6, 2, 1), (5, 3, 1), (4, 3, 2).” &
“So, the possible triples … So, \textbf{three triples in total} … \textit{Wait, hold on, let me check if there are more} … \textit{So, total three triples.}” \\
\addlinespace
\textbf{Conciseness of language} &
“Total N = 2016. Therefore, the remainder is 16.” &
“\textit{Wait, hold on a second.} … \textit{Maybe I can think of all possible partitions} … \textit{No, I think the only possible triples are the three we found.}” \\
\addlinespace
\textbf{Logical signposting} &
“\textbf{Case 1:} s = 1 … \textbf{Case 2:} s = 2 … \textbf{Case 3:} s = 3 (no solutions).” &
“\textbf{Case 1:} S = 1 … \textbf{Subcase 1a} … \textbf{Subcase 1b} … \textit{(digression)} … \textbf{Case 3:} S = 3 … \textit{no solutions} … \textit{(returns to earlier cases).}” \\
\addlinespace
\textbf{Error-checking} &
“Only three possible triples, so the computation is complete.” &
“\textit{Wait, hold on a second. Is that all?} … \textit{let me verify the triples again} … \textit{maybe there are other triples?}” \\
\addlinespace
\textbf{Length} &
Entire solution $\approx$ 200 words. &
Entire solution $\approx$ 370 words (many repeated sentences such as “So, I think 16 is the answer.”). \\
\bottomrule
\end{tabular}
\caption{Qualitative comparison of the shortest correct rollouts from GRPO+L and Deepseek-7B for AIME 25 I, Problem 3. Italicized text in the Deepseek-7B column represents meta-commentary or self-correction loops.}
\label{tab:qualitative_comparison}

\end{table*}

\section{Qualitative Analysis of Solution Conciseness}
\label{sec:appendix_qualitative}

To provide a qualitative illustration of how the length reward enhances conciseness, we contrast the shortest correct solutions generated by GRPO+L and the baseline Deepseek-7B for the same problem (Problem 3, AIME 25 I). Table~\ref{tab:qualitative_comparison} breaks down the comparison across key aspects of readability and reasoning structure.

As the comparison highlights, the GRPO+L model produces a tight, step-by-step solution that remains focused, avoids repetition, and concludes efficiently. In contrast, the Deepseek-7B baseline's reasoning path is less direct, characterized by repeated self-checks and conversational digressions that nearly double the total length and reduce clarity. This case study demonstrates that our length-reward mechanism successfully encourages a more disciplined and economical reasoning style.
\end{document}